# Generating Adversarial Examples With Conditional Generative Adversarial Net

Ping Yu, Kaitao Song, Jianfeng Lu
School of Computer Science and Engineering
Nanjing University of Science and Technology
Nanjing, China
Email: { pingyu, kt.song, lujf }@njust.edu.cn

*Abstract*—Recently, deep neural networks have significant progress and successful application in various fields, but they are found vulnerable to attack instances, e.g., adversarial examples. State-of-art attack methods can generate attack images by adding small perturbation to the source image. These attack images can fool the classifier but have little impact to human. Therefore, such attack instances are difficult to generate by searching the feature space. How to design an effective and robust generating method has become a spotlight. Inspired by adversarial examples, we propose two novel generative models to produce adaptive attack instances directly, in which conditional generative adversarial network is adopted and distinctive strategy is designed for training. Compared with the common method, such as Fast Gradient Sign Method, our models can reduce the generating cost and improve robustness and has about one fifth running time for producing attack instance.

*Keywords—DNN attack; adversarial examples; generative adversarial network*

## I. Introduction

Past several years have observed significant progress of machine learning in many artificial intelligence tasks, including computer vision [1, 2], natural language processing [3]. However, recently researches [4] have made a discovery that most of machine learning models, especially deep learning, are highly vulnerable to some well-designed attack methods. Such attack may cause serious problem, because machine learning has been used for safety-related application, e.g., self-driving, user verification and energy management.

With *universal approximation theorem* [5], researchers achieved great success when they used deep neural network to approximate the ideal function that maps input images to their labels. However, there still exist some differences between the network and the ideal function. More formally, we can define the input as *x*, the classifier as *f*, the label as *y*, and the subspace $\{x|\ f(x) \neq y\}$ as *R*. For most of cases, the subspace *R* definitely exists, which guarantees the possibility of network attack.

The methods to generate *adversarial examples* are the state-of-art attack methods. Adversarial examples refer to those generated instances with small perturbation on source ones. Szegedy et al. [6] discovered that adversarial examples can fool some of start-of-the-art DNN-based models with high probability, such as ResNet [7] and DenseNet [8]. This phenomenon suggests that it is necessary to analyze the cause of adversarial examples and prevent our model from the attack of these examples.

Researchers have proposed various methods to generate adversarial example, such as gradient based methods, Fast Gradient Sign Method(FGSM) [6], search samples based on the gradient of the target model. Optimization-based methods, e.g. Jacobian-based Saliency Map Attack [9], introduce metrics to measure the effect of perturbation and the perturbation are optimized based on this metric.

However, it is expensive for current methods to generate adversarial examples since it costs large computation to search for adversarial example in high-dimension subspace. Besides, we not only should fool the model to produce misclassified outputs, but also should make the difference between adversarial sample and original data as small as possible.

Some other methods find their ways to fool the target with well-designed noise [10]. Vaidya et al. [11] proposed to generate an obfuscated command that is accepted by the victim's speech recognition system but is indecipherable by a human listener. These methods make it possible to access devices like cell phones without authorization.

Most of the previous work either need much computation based on the input, or can be detected by people easily as the instances are obvious artificial noise. In this work, we propose two methods to use the conditional generative adversarial net framework to train a generative model, which can output images able to fool the classifier while remaining correctly classified by a human observer. The first method named conditional generative adversarial network for fake examples (CGAN-F) tries to generate images from Gaussian noise and proves the possibility to generate attack instances directly without source images. We expect that more people can pay attention to the safety of the DNN by the first method. Inspired by the adversarial examples, we propose another method, which is named as conditional generative network for adversarial example (CGAN-Adv). In this method, we separate the generating process into two steps, the generation of raw image and the generation of corresponding perturbation. Then we apply element-wise sum and normalization with the two parts to get the attack image. The CGAN-Adv is efficient in generating images, and it can be observed that the output of CGAN-Adv is better than that of CGAN-F.

The main contributions of our work are summarized as follow:

- Distinguished from the previous work, our methods generate the attack images from noise. In this way, the large storage cost can be saved in DNN attack.
- Most of existing methods require large number of computations to produce an attack example, but our methods only need one round of forward propagation, which are quite efficient.
- We apply normalization after the element-wise sum to address the problem of pixel value overflow.

The rest of paper is organized as follows. Section 2 will review some related work about adversarial examples and generative adversarial networks. In section 3, we give the details of the proposed method. Section 4 gives the experimental result of our methods. Finally, we conclude our work in Section 5.

## II. RELATED WORK

### A. Adversarial examples

The traditional adversarial examples generating methods are widely used in the current state-of-art network attack works. We define $f$ as the classifier, $x$ as raw input and $d_x$ as the desired perturbation. So, we can use a approximate function to model the task of generating adversarial examples as follows

$$\arg\min_{d_x} \|d_x\|, \text{ subject to } f(x+d_x) \neq f(x) \quad (1)$$

Mainly, the methods of generating adversarial examples can be categorized into two types, gradient-based ones and optimization-based ones. The gradient-based methods employ the gradient policy to seek for attacking examples, e.g., Seyed-Mohsen [12] proposed the Deepfool method to find the closet projection of input $x$ on the decision surface. By moving $x$ along the direction of the found projection, it produces perturbation with significantly small magnitude. The advantage of such methods is the high attacking success rate resulting from the full access to the target classifier. The disadvantage is that it is hard to use them to perform black-box attack. The other ones optimize the well-designed metrics to generate adversarial examples, e.g., Nicolas et al. [13] proposed to train a local model to substitute the target DNN. The knowledge of the model and training data for victim is no longer needed. They achieved 96.19% success rate against a well-trained classifier.

More threateningly, empirical evidence shows that adversarial examples often transfer between models trained for the same task [6]. Goodfellow [14] illustrated this phenomenon by introducing the transferability of the adversarial examples. Further researches manifest that adversarial examples occur in large, contiguous regions, or the adversarial subspaces. Florian et al. [15] proposed methods to directly estimate the dimensionality of these subspaces and raise concern about the safety of modern machine learning methods.

### B. Generative adversarial network

Conventionally, generating problems can be modeled as finding a mapping between an input following a given distribution and an output following the target distribution. The key problem is that modeling the target distribution could be extremely hard and time consuming. Recently, Goodfellow et al. [16] introduced the generative adversarial network framework to address this problem. Inspired by the game theory, they trained two models, the generator D – the function to distinguish artificial data from natural data and the discriminator G – the function to generate data out of input following a source distribution aiming at fooling the discriminator. The generator can efficiently generate images. Despite the sound output, the training progress is unstable. By adopting new distance and modification in training strategy, Martin et al. [17] proposed the WGAN algorithm to improve stability of learning strategy and solved the problems like mode collapse.

However, as an unconditioned generative model, the traditional GAN framework does not have the control over the generated data. In other words, we cannot predict the labels of the generated images. Mehdi et al. [18] introduced the conditional generative adversarial net (CGAN) to enroll the label information in the training progress and train a generator producing data of the given class. The infoGAN [19] and ACGAN [20] can also control the output with additional input but are more complex in their framework. The detail of our methods will be introduced in the next section.

## III. APPROACH OVERVIEW

In this section, we will introduce two proposed methods, which can enroll the label information and train generative model at the same time.

### A. CGAN-F

CGAN controls the network output by enrolling extra information $y$ in both training schedule of discriminator and generator. Extra information $y$ can be any kinds of intended information, e.g., class labels or data distribution, and commonly, we feed it into both the generative model and discriminative model as additional input layer.

The overall architecture of our proposed conditional generative adversarial nets for fake images is shown in *Fig. 1*. We define $k$ as the index of source class and $k'$ as the index of the target class. On the side of generative model, the prior input noise $z$ and $y$ are combined to form the conditional input and outputs images $x'$ are generated. From the aspect of the discriminative model, the natural data $x$, or artificial data $x'$, and $y$ are feed into the discriminative model to distinguish the fake examples from the natural data.

We define the adversarial loss function as:

$$L_{CGAN} = E_{x \sim P_{data}(x)}\left[\log D(x|y)\right] + \\ E_{z \sim P_z(z)}\left[\log\left(1-D(G(z|y))\right)\right] \quad (2)$$

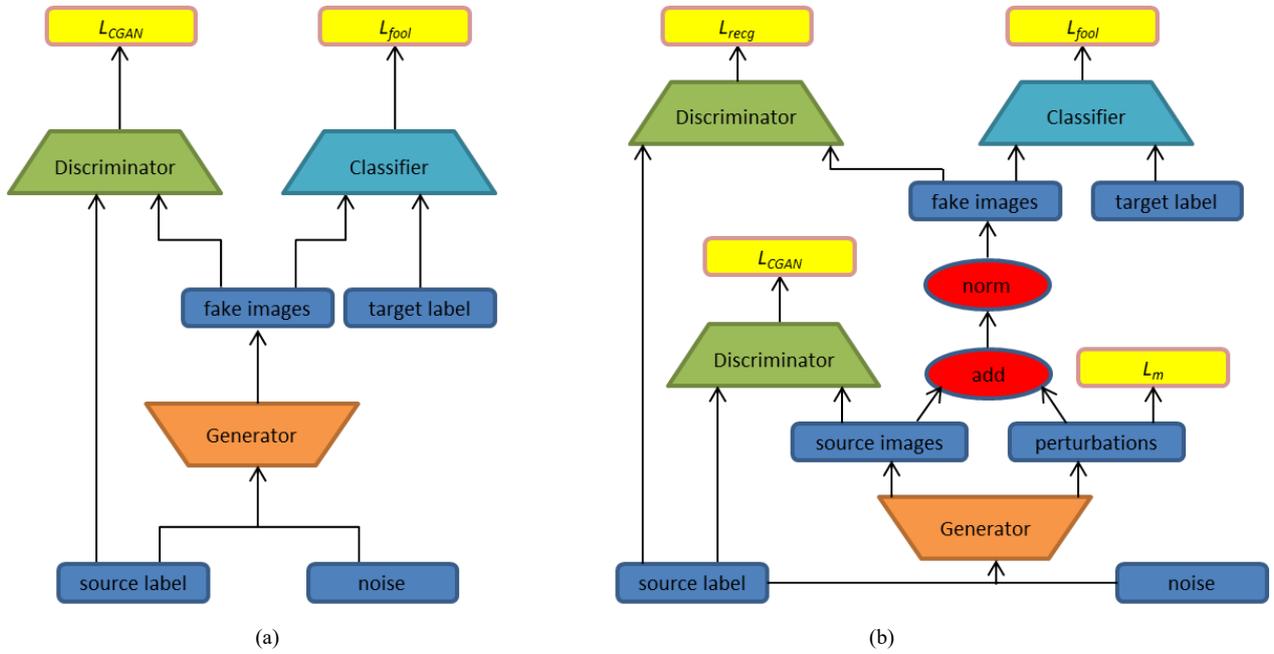

Fig. 1. The overall architecture of our proposed methods. Visualization of the information flow of CGAN-F is in (a), and CGAN-adv in (b).

We add a loss by considering the influence of target to fool the classifier $f$, which can be formulated as:

$$L_{fool} = E\left[loss_f\left(x',k'\right)\right] \quad (3)$$

where $loss_f$ denotes the loss function for training the classifier f. The loss $L_{fool}$ encourages the generator to generate images that are more likely to be misclassified.

Our objective function is described as follows:

$$L = L_{CGAN} + \alpha L_{fool} \quad (4)$$

Where $\alpha$ is a hyper-parameter. Increasing $\alpha$ encourages the generator to output images more likely to be misclassified but may also make the generator produce low quality output.

The traditional training strategy will raise the problem of generating low quality images. To avoid this issue, we make three improvements on the traditional training strategy:

- Pre-train the model with the traditional CGAN training strategy and tune it finely, which can help the model to output sound images of the given class.

- Transform the extra information $y$ from label $k$ to the target label $k'$. It helps the generator to cheat the *game* as the $loss_f$ helps the discriminative model to distinguish between raw and generated ones but generative model gets no *help*.

- After every $n$ iteration, the training process rolls back to the traditional training for one iteration. With this, the generated images are more likely to remain correct classification by a human observer

The complete training algorithm of our model is described in table I.

TABLE I. CGAN-F TRAINING ALGRORITHM

**Algorithm 1**

**Input parameters:** loss weight $\alpha$, batch size $m$, the number of iterations between the traditional training iterations $n$, the total training epoches $N$, the numbers of iteration in each epoch $k$, the loss function $L$, the given class $t$, the target class $t'$.
1: Load the pretrained model.
2: **for** $i = 0 \rightarrow N$ **do**
3:   **for** $j = 0 \rightarrow k$ **do**
4:     Sample images $\{x^{(i)}\}_{i=1\sim m}$ from the real data.
5:     Sample noise $\{z^{(i)}\}_{i=1\sim m}$ from normal distribution.
6:     $x' \leftarrow G(z, t)$
7:     **if** $j$ mod $n$ = 0 **then**
8:       d_fake $\leftarrow D(x', t)$
9:     **else**
10:       d_fake $\leftarrow D(x', t')$
11:     **end if**
12:     d_real $\leftarrow D(x, t)$
13:     d_loss $\leftarrow L(1, $ d_real$)+L(0, $ d_fake$)$
14:     Optimize D
15:     Sample noise $\{z'^{(i)}\}_{i=1\sim m}$ from normal distribution.
16:     $x' \leftarrow G(z,t)$
17:     g_loss $\leftarrow L(1, D(x'))+ \alpha*L(t', x')$
18:     Optimize G
19:   **end for**
20: **end for**

### B. CGAN-Adv

Generating fake images directly can cause the problem of unstable training. Inspired by adversarial example, we train the network to generate images and the corresponding perturbations at the same time. *Fig. 1* shows the overall architecture of our conditional generative adversarial network for adversarial examples.

Since we use the CGAN framework, (1) is adopted. Also, we feed the output of the encoder within the generative model into a new decoder to get the corresponding perturbation $r$, while the old decoder still outputs the artificial image $x'$. To apply this model to targeted attack, we also introduce loss function as (3).

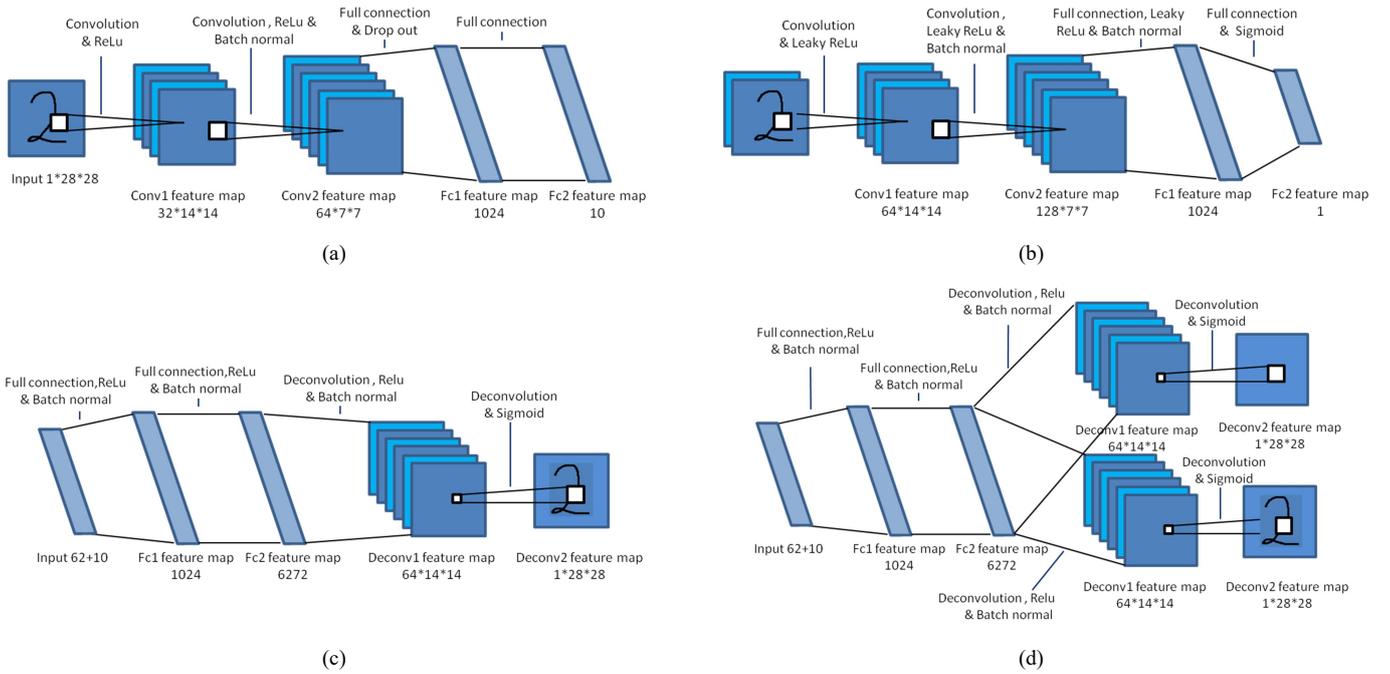

Fig. 2. The architecture of the target classifier is described in (a). The discriminator and generator for CGAN-f are shown in (b) and (c) respectively. The discriminator and generator for CGAN-Adv are shown in (b) and (c) repectively.

The average magnitude of the perturbation is important to make the output similar to the source image. However, if the magnitude becomes too low, the training could become quite unstable. To achieve low magnitude and stabilize the training, we use the following loss:

$$L_m = E\left[mean(abs(r)-c)\right] \qquad (5)$$

where $c$ is the low bound of the average magnitude that we can accept.

Although we tried to add the corresponding perturbations into the source images to get the result, such operation can cause a problem that the pixel value may exceed 255, the upper bound of pixel value. Previous work ignores this problem, and we address it by using normalization technique after element-add operation. The normalization is as follows:

$$x_f = \frac{x' + p*r}{1+p} \qquad (6)$$

where $p$ is a hyper-parameter controlling the weight of the perturbation.

Equation (5) puts a simple constraint on the perturbation, but the perturbation still can wipe out the important part of the generated image. To address this problem, we add a loss function to guarantee that the result is similar to the source class as follows:

$$L_{recg} = E_{z \sim P_z(z)}\left[log(1-D(x_f | y))\right] \qquad (7)$$

Finally, our objective function can be written as:

$$L = L_{CGAN} + \alpha * L_{fool} + \beta * L_m + \gamma * L_{recg} \qquad (8)$$

where $\alpha$, $\beta$ and $\gamma$ are the hyper parameters.

IV. EXPERIMENTAL RESULTS

In this section, we evaluate our methods on the dataset of MNIST [21] and Fashion-MNIST [22]. All experiments are white-box attack. In other words, we have total access to the classifier model.

A. Experimental setup

We use the same model setting for the two classifiers MNIST and Fashion-MNIST. The structure is shown in *Fig. 2*. For convince, we implement the CGAN as described in the Reference [18], so for the generative model, the input vector's dimension should be $d+n$, where $d$ is the dimension of the input noise and $n$ represents the number of classes. In our case, we use $d=62$ and $n=10$. We train them for over 50 epochs and the final performance is 99.1 % and 92.0% on MNIST and Fashion-MNIST respectively.

The corresponding class labels of Fashion-MNIST are t-shirt/top, trouser, pullover, dress, coat, sandal, shirt, sneaker, bag and ankle boot, which map to 0 to 9 respectively.

Both of dataset MNIST and Fashion-MNIST have 10 classes. So, to test the performance on these two datasets, we randomly select two groups of source-target maps as shown in Table II for CGAN-F and CGAN-Adv respectively. In test stage, the input Gaussian noise is same for both CGAN-F

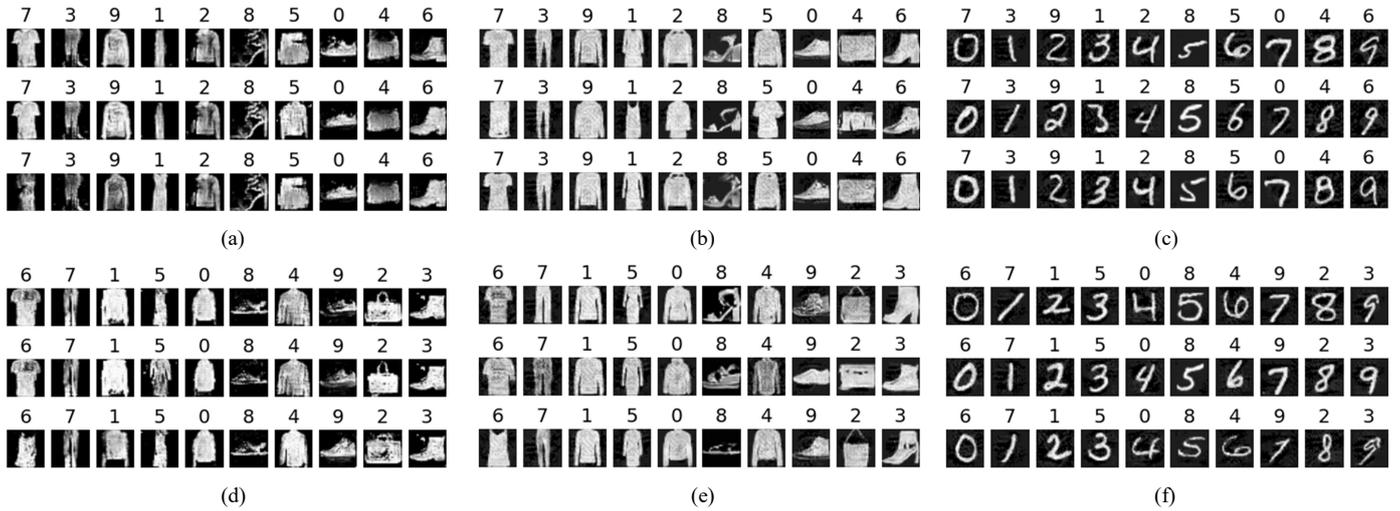

Fig. 3. The results of proposed methods. The source label of the sub-figures in column 0 to 9 is 0 to 9 repectively, and the title of each sub-figure means the classifying result. The results of CGAN-F on Fashion-MNIST for target map1 and 2 are shown in (a) and (d) respectively. The results of CGAN-Adv on Fashion-MNIST are in (b) and (e), and the results on MNIST are in (c) and (f).

and CGAN-Adv.

TABLE II. SOURCE-TO-TARGET MAP OF CLASSES

| Map index | Map |
|---|---|
| 1 | {0:7, 1:3, 2:9, 3:1, 4:2, 5:8, 6:5, 7:0, 8:4, 9:6} |
| 2 | {0:6, 1:7, 2:1, 3:5, 4:0, 5:8, 6:4, 7:9, 8:2, 9:3} |

The parameters for generative model and the discriminative model of CGAN-F is set as follows: learning rate 0.0002, the batch size 128, the number of epochs 25. Some generated images are presented in *Fig. 4*.

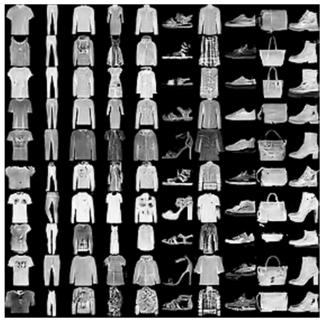

Fig. 4. Output of the pretrained CGAN.

### B. CGAN-F

For generative model, we employ an encoder with two linear layers and a decoder containing two deconvolution layers. For discriminative model, we adopt two convolution layers and two linear layers. The detail of the model is described in *Fig. 2*.

From *Fig. 3*, we can find that CGAN-F can perform well when dealing with source-target pairs that are similar in outline, e.g., pullover and coat. However, when it comes to pairs distinguished in outline, the performance was unsatisfied, e.g., trouser and sneaker. The results indicate that it is possible to generate attack examples directly with GAN.

### C. CGAN-Adv

To verify with CGAN-Adv, we build the generative model with an encoder containing two linear layers and two decoders that have two deconvolution layers respectively. The setting of discriminative model is barely the same as that of CGAN. The detail of the generator is described in *Fig. 2*.

We train our models with Adam-optimizer [23]. The batch size is 64. The initial learning rate is 0.0002 and decrease 0.1 every 10 epochs. The hyper parameters are set as: $\alpha$=0.5, $\beta$=50, $\gamma$=0.4, $p$=0.25 and $c$=0.23. The generated images on Fashion-MNIST are shown in *Fig. 3*.

Besides, the results on MNIST are shown in *Fig. 3(??)*. For most results, human can easily recognize their labels. However, the classifier misclassified all the examples shown in the figure. We adjust our hyper parameters $\beta$ and $c$ to reduce the magnitude of perturbation, but it may affect the success attack rate.

The purpose of introducing loss Lm is to low the average magnitude of perturbation, and we find the average magnitude is only around 13.2% of the raw data and 5% of the maximum magnitude by statistic. The low average magnitude suggests a good similarity between the final images and source images. Also, through parallel training with *Lrecg*, we find the results of CGAN-Adv appear better than that of CGAN-F. It can be seen that some details of the images are preserved, e.g., pattern on the shirts and holes on the sandals. Some of the output images are shown in *Fig. 5*. From the images of perturbation, we can observe that the generative model tries to find the key points in recognizing the target label, and by increasing the values of the key pixels, it can fool the classifier.

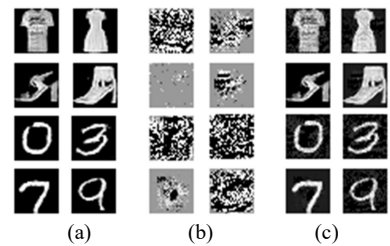

Fig. 5. The generated source images are shown in (a). The corresponding perturbations are shown in (b). The final results are shown in (c).

We further evaluate the efficiency of CGAN-Adv based on the average time for producing a single attack instance. We compare CGAN-Adv with some common methods, such as the iterative least-likely class method (ILCM) [24], the FGSM, the Deepfool and the WRM [25]. We use the five methods to generate 10k attack instances and all the methods achieve high attack success rate. The average runtime is shown in Table III, and it can be observed that CGAN-Adv can output instances much faster than the other two methods. Besides, the traditional adversarial example generating methods require the source image to produce an attack instance, therefore, additional storage space is needed. However, CGAN-Adv does not need additional storage space, which makes CGAN-Adv advantageous when applying to large-scale attack.

TABLE III. AVEARGE TIME FOR PRODUCING AN INSTANCE

| Dataset | MNIST | Fashion-MNIST |
|---|---|---|
| ILCM | 62ms | 65ms |
| FGSM | 53ms | 61ms |
| Deepfool | 59ms | 61ms |
| WRM | 73ms | 80ms |
| CGAN-Adv(ours) | 13ms | 13ms |

## V. CONCLUSION

In this work, we propose two methods to generate images to attack a trained DNN classifier, namely CGAN-F and CGAN-Adv. The CGAN-F can output the attack images in one stage, while the CGAN-Adv in two stages. Our experiments have shown the possibility to generate images directly by using GAN, and such images can be misclassified by the classifier while being correctly classified by human. Compared with previous work, our work can produce attack images more efficiently.

In the future, we will concentrate on find better attack image and reduce the magnitude of the perturbation.

## *Acknowledgment*

This work was funded by the National Key R&D Program of China and contract number is 2017YFB1300205.

## *References*

[1] K. Simonyan and A. Zisserman, "Very deep convolutional networks for large-scale image recognition," arXiv preprint arXiv:1409.1556, 2014.

[2] R. Girshick, "Fast r-cnn," in Proceedings of the IEEE international conference on computer vision, 2015, pp. 1440-1448.

[3] E. Cambria and B. White, "Jumping NLP Curves: A Review of Natural Language Processing Research [Review Article]," IEEE Computational Intelligence Magazine, vol. 9, pp. 48-57, 2014.

[4] C. Szegedy, W. Zaremba, I. Sutskever, J. Bruna, D. Erhan, I. Goodfellow, et al., "Intriguing properties of neural networks," arXiv preprint arXiv:1312.6199, 2013.

[5] K. Hornik, M. Stinchcombe, and H. White, "Multilayer feedforward networks are universal approximators," Neural Networks, vol. 2, pp. 359-366, 1989.

[6] I. J. Goodfellow, J. Shlens, C. Szegedy, I. J. Goodfellow, J. Shlens, and C. Szegedy, "Explaining and harnessing adversarial examples," in ICML, 2015, pp. 1-10.

[7] K. He, X. Zhang, S. Ren, and J. Sun, "Deep Residual Learning for Image Recognition," in 2016 IEEE Conference on Computer Vision and Pattern Recognition (CVPR), 2016, pp. 770-778.

[8] G. Huang, Z. Liu, L. v. d. Maaten, and K. Q. Weinberger, "Densely Connected Convolutional Networks," in 2017 IEEE Conference on Computer Vision and Pattern Recognition (CVPR), 2017, pp. 2261-2269.

[9] N. Papernot, P. McDaniel, S. Jha, M. Fredrikson, Z. B. Celik, and A. Swami, "The Limitations of Deep Learning in Adversarial Settings," in 2016 IEEE European Symposium on Security and Privacy (EuroS&P), 2016, pp. 372-387.

[10] N. Carlini, P. Mishra, T. Vaidya, Y. Zhang, M. Sherr, C. Shields, et al., "Hidden Voice Commands," in USENIX Security Symposium, 2016, pp. 513-530.

[11] T. Vaidya, Y. Zhang, M. Sherr, and C. Shields, "Cocaine noodles: exploiting the gap between human and machine speech recognition," in Usenix Conference on Offensive Technologies, 2015, pp. 16-16.

[12] S.-M. Moosavi-Dezfooli, A. Fawzi, and P. Frossard, "Deepfool: a simple and accurate method to fool deep neural networks," in Proceedings of the IEEE Conference on Computer Vision and Pattern Recognition, 2016, pp. 2574-2582.

[13] N. Papernot, P. McDaniel, I. Goodfellow, S. Jha, Z. B. Celik, and A. Swami, "Practical black-box attacks against machine learning," in Proceedings of the 2017 ACM on Asia Conference on Computer and Communications Security, 2017, pp. 506-519.

[14] N. Papernot, P. McDaniel, and I. Goodfellow, "Transferability in machine learning: from phenomena to black-box attacks using adversarial samples," arXiv preprint arXiv:1605.07277, 2016.

[15] F. Tramèr, N. Papernot, I. Goodfellow, D. Boneh, and P. McDaniel, "The Space of Transferable Adversarial Examples," arXiv preprint arXiv:1704.03453, 2017.

[16] I. Goodfellow, J. Pouget-Abadie, M. Mirza, B. Xu, D. Warde-Farley, S. Ozair, et al., "Generative adversarial nets," in Advances in neural information processing systems, 2014, pp. 2672-2680.

[17] M. Arjovsky, S. Chintala, and L. Bottou, "Wasserstein generative adversarial networks," in International Conference on Machine Learning, 2017, pp. 214-223.

[18] M. Mirza and S. Osindero, "Conditional generative adversarial nets," arXiv preprint arXiv:1411.1784, 2014.

[19] X. Chen, Y. Duan, R. Houthooft, J. Schulman, I. Sutskever, and P. Abbeel, "Infogan: Interpretable representation learning by information maximizing generative adversarial nets," in Advances in Neural Information Processing Systems, 2016, pp. 2172-2180.

[20] A. Odena, C. Olah, and J. Shlens, "Conditional image synthesis with auxiliary classifier gans," arXiv preprint arXiv:1610.09585, 2016.

[21] Y. Lecun and C. Cortes, "The mnist database of handwritten digits," 2010.

[22] H. Xiao, K. Rasul, and R. Vollgraf, "Fashion-MNIST: a Novel Image Dataset for Benchmarking Machine Learning Algorithms," 2017.

[23] D. P. Kingma and J. Ba, "Adam: A Method for Stochastic Optimization," Computer Science, 2014.

[24] Kurakin, Alexey, Ian Goodfellow, and Samy Bengio, "Adversarial examples in the physical world," arXiv preprint arXiv:1607.02533, 2016.

[25] Sinha, Aman, Hongseok Namkoong, and John Duchi, "Certifiable Distributional Robustness with Principled Adversarial Training," arXiv preprint arXiv:1710.10571, 2017.